\documentclass{article}
\usepackage{spconf,amsmath,amssymb,graphicx,booktabs,multirow}
\usepackage[T1]{fontenc}
\usepackage[utf8]{inputenc}
\usepackage{CJKutf8}
\usepackage{url}
\usepackage[table]{xcolor}
\usepackage[hidelinks]{hyperref}

\definecolor{metricgray}{gray}{0.84}
\definecolor{todored}{RGB}{155,35,35}
\newcommand{\best}[1]{\textcolor{red}{\textbf{#1}}}

\newcommand{\X}{\mathbf{X}}
\newcommand{\Oseq}{\mathbf{O}}
\newcommand{\Rseq}{\mathbf{R}}
\newcommand{\Ha}{\mathbf{H}^{\mathrm a}}
\newcommand{\Hm}{\mathbf{H}^{\mathrm m}}
\newcommand{\Za}{\mathbf{Z}^{\mathrm a}}
\newcommand{\jp}[1]{\begin{CJK*}{UTF8}{min}#1\end{CJK*}}

\title{Ruby-ASR: Evidence-Preserving Supervision for Joint Orthographic and Lexical-Reading Recognition}

\name{
\shortstack{
Hao Shi, Yun Liu, Xuehao Yang, \\ Jun Liu, Chuanbo Hua, Xuanjun Chen, Lianbo Liu, Shiao Zhu, Zixiong Su
}}
\address{
Independent Researcher \\
\begin{tabular}{c}
\texttt{hshi@ieee.org,liuyun5620472@gmail.com,sunskyxh@gmail.com} \\
\texttt{liujun@alu.uestc.edu.cn,cbhua@kaist.ac.kr,zxsu@g.ecc.u-tokyo.ac.jp}
\end{tabular}
}

\begin{document}
\ninept
\maketitle

% -------------------------------------------------------------------------
\begin{abstract}
Conventional Japanese automatic speech recognition (ASR) is supervised by an
orthographic transcript, although the same written form can correspond to
different lexical readings realized in speech.  Such utterances receive an
identical target, so their reading distinction is absent from the supervision
interface and cannot be recovered reliably by post-hoc text-only
grapheme-to-phoneme conversion.  We present \textbf{Ruby-ASR}, which refines
the conventional target into a span-bound orthographic--lexical-reading
sequence, e.g., \jp{今日[きょう]}.  Unlike separate full-sentence orthographic
and phonological outputs, the ruby representation locally binds each written
span to its realized reading and permits deterministic recovery of both views.
We instantiate the target under subtitle-style and verbatim-style
transcription conventions using a Qwen3-ASR backbone; a mora-level CTC
objective provides auxiliary monotonic reading supervision.  
The experimental results across five
Japanese benchmarks
%, Ruby-ASR-ver reduces character-count-weighted Kana CER
%from 5.74\% for Qwen3-ASR and 5.64\% for ReazonSpeech-k2 to 3.84\%, while
%Ruby-ASR-sub obtains weighted Raw CER and script-aware CER of 8.51\% and
%6.64\%, respectively.  These results 
show that refining the recognition target
can improve lexical-reading recovery without sacrificing readable
orthographic transcription.  We release the checkpoints and inference code.
\end{abstract}


\section{Introduction}
\label{sec:intro}
Automatic speech recognition is conventionally formulated as mapping speech
to an orthographic transcription. For Japanese, the output is a readable
mixture of kanji and kana and is commonly evaluated using character error rate
(CER)~\cite{karita21_interspeech,karita2023lenient}. However, it does not
always identify how a lexical item was realized in speech. Japanese
orthography and lexical reading are not in one-to-one correspondence: different
written forms can share a reading, while one written form can admit multiple
readings~\cite{sudo23b_interspeech,zhang2023pronunciation,
sato2022pronunciation_corpora,liu2026sarashina22tts}.
If two utterances have different readings but the same orthographic transcript,
orthographic ASR assigns them an identical target and does not require this
distinction at the output. We call this \emph{target-level lexical-reading
collapse}. A downstream grapheme-to-phoneme (G2P) model cannot resolve the
problem because it observes only the recognized text and infers a likely
reading from linguistic context
\cite{matogawa2024japanese,passaglia2023yomikata,
koriyama2026japanese_g2p}. Likewise, text-only kanji-reading benchmarks
\cite{liu2026sarashina22tts,mibayashi2026yomi_bench} cannot test whether a
reading follows the input speech. Conversely, kana- or phoneme-only ASR
preserves pronunciation but loses the orthographic distinctions and
readability provided by kanji.

Joint phoneme--grapheme ASR has addressed this tradeoff using shared encoders
and separate utterance-level decoders
\cite{kubo2020joint_phoneme_grapheme,shi2026persistent,nadig2020multitarget,
krishna2021dual_decoder}. More closely, Omachi et al. serialize each grapheme
or multi-grapheme span with its aligned phonemic annotation in a single
sequence~\cite{omachi2021joint_annotations}. Transcript-prompted Whisper
predicts Japanese phonemic and prosodic annotations from audio conditioned on
a ground-truth transcript~\cite{hu2025transcript_prompted}, but is an
annotation model rather than speech-only ASR. The remaining challenge is to
scale speech-conditioned lexical-reading supervision while retaining a
directly usable orthographic transcript.

We therefore propose \textbf{Ruby-ASR}, which replaces the conventional target
with a structured ruby-form sequence such as ``\jp{今日[きょう]}.'' This
refines the output equivalence classes induced by orthographic supervision:
utterances with the same surface form but different realized readings receive
different labels. Its local span--reading binding permits deterministic
recovery of both conventional orthographic and lexical-reading transcripts.
We train two variants with the same architecture and objective:
\emph{Ruby-ASR-sub} follows a concise subtitle-style convention, whereas
\emph{Ruby-ASR-ver} follows a more literal verbatim-style convention. Ruby
generation is the primary inference path, while a shared mora-level CTC branch
provides auxiliary monotonic supervision and can optionally produce a
decoder-free reading hypothesis.

Our contributions are threefold. First, we formulate lexical-reading loss as
a target-level collapse under orthographic supervision and introduce an
oracle-orthography G2P diagnostic that isolates the empirical text-only
reconstruction gap from orthographic ASR errors. Second, we scale span-bound
ruby supervision to a large LLM-based Japanese ASR system using an
evidence-constrained target-construction pipeline. Third, evaluation on five
Japanese benchmarks shows improved direct lexical-reading recognition with
competitive orthographic accuracy; the checkpoints and inference code are
publicly released\footnote{\url{https://github.com/hshi-speech/Ruby-ASR-1.7B}}\footnote{\url{https://huggingface.co/hshispeech/Ruby-ASR-1.7B}}.

% -------------------------------------------------------------------------
\section{Beyond Orthographic Supervision}
\label{sec:problem}

\subsection{Target-level lexical-reading collapse}
\vspace{-5pt}

Let $\X$ denote an utterance, $\Oseq$ its orthographic representation, and
$\Rseq$ its lexical reading realized in speech. Conventional Japanese ASR
learns
\begin{equation}
	\setlength{\abovedisplayskip}{0pt}
	\setlength{\belowdisplayskip}{0pt}
    p(\Oseq\mid\X).
    \label{eq:orth_asr}
\end{equation}
For lexical reading, orthography is not always sufficient:
\begin{equation}
	\setlength{\abovedisplayskip}{0pt}
	\setlength{\belowdisplayskip}{0pt}
    H(\Rseq\mid\Oseq)>0,
    \qquad
    I(\Rseq;\X\mid\Oseq)>0.
    \label{eq:residual_info}
\end{equation}
For $\Oseq_a=\Oseq_b$ but $\Rseq_a\ne\Rseq_b$, orthographic supervision maps
$\X_a$ and $\X_b$ to the same target, so the reading distinction is absent
from the learning interface. Equivalently, it induces
\begin{equation}
	\setlength{\abovedisplayskip}{0pt}
	\setlength{\belowdisplayskip}{0pt}
    \X_a\sim_{\mathrm O}\X_b
    \iff \Oseq_a=\Oseq_b,
    \label{eq:orth_equivalence}
\end{equation}
whereas Ruby supervision refines the relation to
\begin{equation}
	\setlength{\abovedisplayskip}{0pt}
	\setlength{\belowdisplayskip}{0pt}
    \X_a\sim_{\mathrm{ruby}}\X_b
    \iff (\Oseq_a,\Rseq_a)=(\Oseq_b,\Rseq_b),
    \label{eq:ruby_equivalence}
\end{equation}
thereby separating examples collapsed by orthographic supervision.

\vspace{-5pt}

\subsection{Supervision interfaces and prior joint targets}
\vspace{-5pt}

Table~\ref{tab:interfaces} compares the relevant supervision interfaces.
Orthographic ASR preserves readability but omits the realized reading; ASR
followed by G2P infers it only after speech has been reduced to text. Kana-only
ASR directly recognizes reading but loses orthographic identity. Prior
multi-target systems retain both views using separate utterance-level sequence
heads~\cite{kubo2020joint_phoneme_grapheme,nadig2020multitarget,
krishna2021dual_decoder}, but do not explicitly bind their corresponding
spans. Omachi et al. instead combine locally aligned grapheme, phoneme, and
optional POS subsequences in one autoregressive target
\cite{omachi2021joint_annotations}. Ruby-ASR scales this closely related
interface to LLM-based Japanese ASR as
\jp{今日[きょう]学校[がっこう]に行[い]く}, with deterministic orthographic and
reading projections. In contrast, transcript-prompted annotation requires
ground-truth orthography at inference time~\cite{hu2025transcript_prompted}.

\vspace{-5pt}

% Keep Table~\ref{tab:interfaces} unchanged here.

\subsection{Text-only reconstruction gap}
\vspace{-5pt}

For recognized orthography, post-hoc G2P predicts
\begin{equation}
	\setlength{\abovedisplayskip}{0pt}
	\setlength{\belowdisplayskip}{0pt}
    \widehat{\Rseq}_{\mathrm{ASR+G2P}}
    =\mathrm{G2P}(\widehat{\Oseq}).
    \label{eq:asr_g2p}
\end{equation}
This conflates orthographic recognition errors with text-only reading
reconstruction. We therefore also evaluate
\begin{equation}
	\setlength{\abovedisplayskip}{0pt}
	\setlength{\belowdisplayskip}{0pt}
    \widehat{\Rseq}_{\mathrm{oracle\text{-}G2P}}
    =\mathrm{G2P}(\Oseq^*),
    \label{eq:oracle_g2p}
\end{equation}
where $\Oseq^*$ is ground-truth orthography, and define
\begin{equation}
	\setlength{\abovedisplayskip}{0pt}
	\setlength{\belowdisplayskip}{0pt}
    E_{\mathrm{text}}
    =\mathrm{CER}\!\left(\mathrm{G2P}(\Oseq^*),\Rseq^*\right)
    \label{eq:text_gap}
\end{equation}
as the \emph{empirical text-only reconstruction gap} of the selected G2P
system, not a theoretical irreducible bound. To avoid circularity, the primary
analysis uses only reading references that are manually verified or prepared
independently of that G2P system.

\begin{table}[t]
\centering
\footnotesize
\setlength{\tabcolsep}{2.8pt}
\renewcommand{\arraystretch}{1.08}

\caption{Comparison of orthography--reading supervision interfaces.
``Orth.'' and ``Reading'' denote supervised outputs; ``Span-bound'' denotes
explicit correspondence between each $O_j$ and $R_j$; and ``Speech for $R$''
denotes speech-conditioned reading prediction rather than post-ASR G2P.
Separate sequence targets lack explicit span correspondence, whereas Ruby-ASR
uses an interleaved span-bound target.}
\label{tab:interfaces}

\vspace{3pt}
\begin{tabular}{@{}lcccc@{}}
\toprule
\textbf{Interface}
& \textbf{Orth.}
& \textbf{Reading}
& \shortstack{\textbf{Span-}\\\textbf{bound}}
& \shortstack{\textbf{Speech}\\\textbf{for} $\boldsymbol{R}$} \\
\midrule

Orthographic ASR
& \checkmark
& --
& --
& N/A \\

ASR + G2P
& \checkmark
& Inferred
& --
& No \\

Kana/phoneme ASR
& --
& \checkmark
& --
& Yes \\

Separate sequence heads
& \checkmark
& \checkmark
& --
& Yes \\

Interleaved span target
& \checkmark
& \checkmark
& \checkmark
& Yes \\

\bottomrule
\end{tabular}
\vspace{-10pt}
\end{table}

\section{Ruby-ASR}
\label{sec:method}

\subsection{Speech--language backbone}
\vspace{-5pt}

Ruby-ASR is initialized from Qwen3-ASR~\cite{shi2026qwen3asr}.  Given speech
$\X$, the acoustic encoder and modality projector produce
\begin{equation}
	\setlength{\abovedisplayskip}{0pt}
	\setlength{\belowdisplayskip}{0pt}
    \Ha=\mathcal E_{\mathrm{aud}}(\X),
    \qquad
    \Za=\mathcal P(\Ha).
    \label{eq:backbone}
\end{equation}
An autoregressive language-model decoder predicts
\begin{equation}
	\setlength{\abovedisplayskip}{0pt}
	\setlength{\belowdisplayskip}{0pt}
    p(\mathbf Y\mid\X)
    =\prod_{n=1}^{N}p(y_n\mid\mathbf Y_{<n},\Za).
    \label{eq:ar}
\end{equation}
Under conventional supervision, $\mathbf Y=\Oseq$ and the realized reading is
not part of the output target.

\subsection{Span-bound Ruby target}
\vspace{-5pt}
\label{sec:ruby_target}

Let $O_j$ denote an annotated orthographic span, $R_j$ its realized reading,
and $U_j$ unannotated material whose reading is already explicit, primarily
kana.  We serialize the Ruby target as
\begin{equation}
	\setlength{\abovedisplayskip}{0pt}
	\setlength{\belowdisplayskip}{0pt}
\mathbf Y^{\mathrm{ruby}}
=
U_0O_1[R_1]U_1\cdots O_J[R_J]U_J.
\label{eq:ruby_target}
\end{equation}
Readings are normalized to hiragana, and brackets are reserved structural
tokens.  Annotation is performed at lexical-span rather than character level,
because compound readings are not necessarily compositional.  For example,
\jp{今日[きょう]} forms one pair, while okurigana remains outside the
annotation, as in \jp{行[い]く}.  A complete target may therefore be
\jp{明日[あした]は学校[がっこう]に行[い]く}.  This extends the interleaved
transcription--annotation formulation of prior work~\cite{omachi2021joint_annotations}
to Japanese lexical-span Ruby.

Both output views can be recovered deterministically:
\begin{equation}
	\setlength{\abovedisplayskip}{0pt}
	\setlength{\belowdisplayskip}{0pt}
\begin{aligned}
\widehat{\Oseq}
&=\mathcal P_{\mathrm{orth}}
(\widehat{\mathbf Y}^{\mathrm{ruby}})
=U_0O_1U_1\cdots O_JU_J,\\
\widehat{\Rseq}
&=\mathcal P_{\mathrm{read}}
(\widehat{\mathbf Y}^{\mathrm{ruby}})
=\rho(U_0)R_1\rho(U_1)\cdots R_J\rho(U_J),
\end{aligned}
\label{eq:projections}
\end{equation}
where $\rho(\cdot)$ normalizes unannotated material and removes non-spoken
symbols.  Thus, one generated sequence provides locally aligned orthographic
and lexical-reading transcriptions. 
The Ruby sequence is trained autoregressively:
\begin{equation}
	\setlength{\abovedisplayskip}{0pt}
	\setlength{\belowdisplayskip}{0pt}
\mathcal L_{\mathrm{ruby}}
=
-\sum_{n=1}^{N}
\log p\!\left(
y_n^{\mathrm{ruby}}
\mid
\mathbf Y_{<n}^{\mathrm{ruby}},\Za
\right).
\label{eq:ruby_loss}
\end{equation}
This makes the realized reading an explicit speech-conditioned target.
However, because $R_j$ follows $O_j$, the decoder may also exploit preceding
orthographic context; Ruby supervision preserves and binds reading
information, but does not by itself prove causal reliance on speech.

\begin{figure}[t]
    \centering
    \IfFileExists{pics/ruby_asr.pdf}{%
      \includegraphics[width=\linewidth]{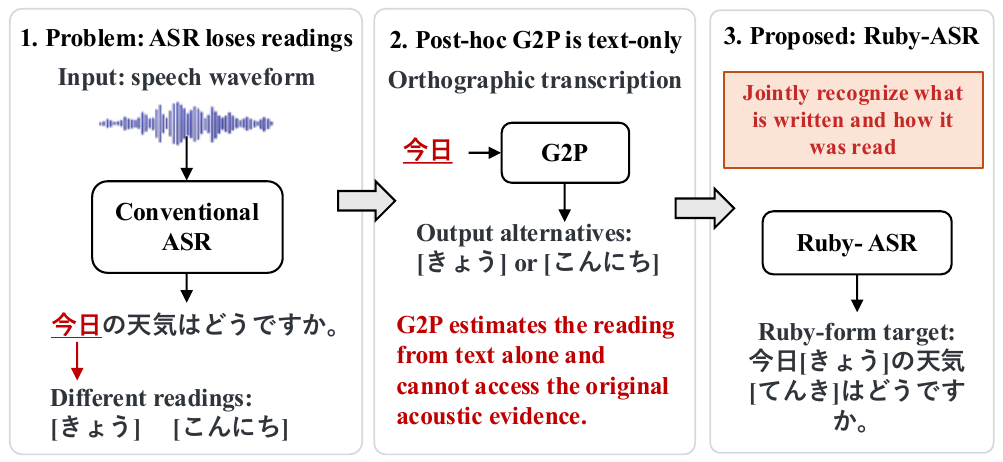}}{%
      \fbox{\parbox[c][34mm][c]{0.94\linewidth}{\centering
      Placeholder for the supervision-collision and Ruby-ASR overview figure.}}}
    \vspace{-20pt}
    \caption{Motivation and overview.  Orthographic supervision assigns an
    identical target to acoustically distinct readings.  Ruby-ASR instead
    binds each orthographic span to the reading realized in speech.}
    \label{fig:overview}
    \vspace{-10pt}
\end{figure}

\subsection{Auxiliary mora-level regularization}
\vspace{-5pt}

To provide complementary monotonic reading supervision, the shared encoder is
followed by additional Transformer layers
\cite{11434801,10889634,10832330,shi2026distillingllmsemanticpriors} and a
mora-level CTC head:
\begin{equation}
	\setlength{\abovedisplayskip}{0pt}
	\setlength{\belowdisplayskip}{0pt}
    \Hm=\mathcal T_{\mathrm{mora}}(\Ha),
    \qquad
    \mathcal L_{\mathrm{CTC}}
    =-\log p_{\mathrm{CTC}}(\mathbf M\mid\Hm),
    \label{eq:ctc}
\end{equation}
where $\mathbf M$ is a mora-level reading sequence.  Joint training uses
\begin{equation}
	\setlength{\abovedisplayskip}{0pt}
	\setlength{\belowdisplayskip}{0pt}
    \mathcal L
    =(1 - \lambda)\mathcal L_{\mathrm{ruby}}
     +\lambda \mathcal L_{\mathrm{CTC}}.
    \label{eq:joint_loss}
\end{equation}
Ruby generation remains the main inference path.  The CTC branch can also
produce an optional decoder-free kana hypothesis, but is treated primarily as
auxiliary regularization in this work.

\subsection{Automatic target construction}
\vspace{-5pt}

Most source corpora contain orthographic transcripts but no speech-grounded
lexical readings.  Pronunciation-annotated text corpora
\cite{sato2022pronunciation_corpora} and furigana-annotated speech resources
\cite{sato2025aozora_speech,ndl2025aozora_corpus} provide valuable coverage,
but do not supply matched annotations for the full large-scale ASR training
pool.  We therefore use evidence-constrained closed-set annotation.  Each
transcript is normalized and morphologically segmented to identify
kanji-bearing lexical units and separate surface okurigana.  Multiple
open-source ASR systems filter unreliable audio--text pairs, and an ensemble
of three CTC models aligns reading evidence to the retained spans.  For each
$O_j$, OpenJTalk supplies a candidate set $\mathcal C(O_j)$; an open-source
language model may resolve contextual ambiguity only by selecting
$R_j\in\mathcal C(O_j)$, rather than freely generating a label.  Nonempty,
canonical-kana candidates consistent with the aligned utterance reading are
serialized as $O_j[R_j]$; unresolved or structurally invalid samples are
excluded from Ruby supervision.

% -------------------------------------------------------------------------
\section{Experimental Setup}
\label{sec:setup}

\subsection{Training data and transcription conventions}
\vspace{-5pt}
We fine-tune Qwen3-ASR using Eq.~\eqref{eq:joint_loss} on a source pool of
approximately 93.1 million audio--text segments (171.5 thousand hours) of
read, spontaneous, expressive, and synthetic Japanese speech. It combines
internally curated data with ReazonSpeech~\cite{yin2023reazonspeech}, Common
Voice 26~\cite{ardila2020commonvoice}, and selected JSUT
subsets~\cite{sonobe2017jsut}. No evaluation audio from B5K, CSJ, Book, CV8,
or TEDx is used for training. Synthetic speech increases coverage of
medium- and high-difficulty lexical readings.
We construct separate training sets for two transcription conventions using
the same architecture and objective. \emph{Ruby-ASR-sub} follows a concise
subtitle-style convention similar to Whisper and ReazonSpeech.
%; its final training set contains approximately 16,000 hours.
\emph{Ruby-ASR-ver} follows Qwen3-ASR's more literal convention and uses a
substantially larger verbatim-style subset. We fine-tune all parameters with
$\lambda=0.3$. The CTC branch predicts 278 mora-level classes
through a two-layer, eight-head Transformer adapter. Training uses
length-grouped dynamic batches with a 600-s padded-audio budget and bfloat16
AdamW with a peak learning rate of $2\times10^{-5}$, cosine decay, and 1\%
linear warm-up. Starting from an earlier fine-tuned checkpoint with a
reinitialized optimizer, we train for one epoch (185,186 steps), evaluate
every 1,000 steps, and select the checkpoint with the lowest development loss.

\subsection{Benchmarks and baselines}
\vspace{-5pt}

We evaluate on JSUT-BASIC5000 (B5K), the Corpus of Spontaneous Japanese (CSJ)
\cite{maekawa2003csj}, JSUT-Book (Book), Common Voice 8 (CV8), and
TEDx~\cite{ando2021construction}.  Baselines are Whisper-Large-v3 (Whisp.)
\cite{radford2023whisper}, Kotoba-v2 (Kotoba)
\cite{gandhi2023distilwhisper}, NeMo- and k2-based ReazonSpeech systems
(Rz-NeMo and Rz-k2), Qwen3-ASR (Qwen)~\cite{shi2026qwen3asr}, and
Kana-Whisper (K-Whisp.)~\cite{liu2026sarashina22tts}.

\subsection{Metrics}
\vspace{-5pt}

References and hypotheses are consistently normalized for numerals, long
vowels, sokuon, punctuation, ruby marks, and kana variants. \textbf{Raw CER}
evaluates script-preserving orthographic transcription, whereas \textbf{SA-CER}
compares kana regions by reading and kanji regions by orthography, reducing
penalties from kana--kanji variation. \textbf{Kana CER} evaluates lexical
readings using Ruby-ASR's direct reading projection or a shared G2P for
orthographic baselines. Oracle G2P applies the same system to ground-truth
orthography. To avoid circularity, primary oracle results are limited to B5K
and CSJ, whose reading references were independently verified.
For correctly recognized orthographic spans, we define
\begin{equation}
\setlength{\abovedisplayskip}{0pt}
\setlength{\belowdisplayskip}{0pt}
\begin{aligned}
\mathcal J_{\mathrm{corr}}
={j:\widehat O_j=O_j^*},\
\mathrm{C\text{-}KanaCER}
=\frac{\sum_{j\in\mathcal J_{\mathrm{corr}}}
d_{\mathrm{edit}}(\widehat R_j,R_j^*)}
{\sum_{j\in\mathcal J_{\mathrm{corr}}}|R_j^*|}.
\end{aligned}
\label{eq:conditional_cer}
\end{equation}
We report it for all eligible spans and the multi-reading subset
$|\mathcal C(O_j^*)|>1$, together with span-level exact-reading accuracy
($\widehat R_j=R_j^*$). Rare and unseen subsets are defined by the training
frequency $f(O_j^*,R_j^*)$ of each exact orthography--reading pair:
$1\leq f<10$ and $f=0$, respectively; evaluation data are excluded from these
counts.

\begin{table}[t]
\centering
\scriptsize
\setlength{\tabcolsep}{2.6pt}
\renewcommand{\arraystretch}{0.94}
\caption{Raw CER and script-aware CER (SA-CER, \%) on five benchmarks.
W.Avg. is character-count weighted. Lower is better.}
\label{tab:orth_results}
\begin{tabular}{@{}lcccccc@{}}
\toprule
Model & B5K & CSJ & Book & CV8 & TEDx & W.Avg. \\
\midrule
\multicolumn{7}{c}{\cellcolor{metricgray}\textbf{Raw CER}}\\
Whisp.       & 7.20 & 19.77 & \best{20.28} & 8.27 & 9.52 & 11.36 \\
Kotoba       & 8.98 & 18.95 & 22.63 & 9.39 & 10.75 & 12.33 \\
Rz-NeMo      & 7.38 & 19.24 & 20.75 & 8.78 & 10.66 & 11.79 \\
Rz-k2        & \best{6.52} & 16.82 & 20.73 & 7.64 & 9.09 & 10.35 \\
Qwen         & 8.71 & 14.55 & 23.70 & 9.18 & 9.08 & 10.78 \\
\midrule
Ruby-ASR-ver & 8.23 & 8.23 & 23.50 & 8.12 & \best{8.90} & 9.10 \\
Ruby-ASR-sub & 7.84 & \best{6.33} & 23.21 & \best{7.25} & 9.06 & \best{8.51} \\
\midrule
\multicolumn{7}{c}{\cellcolor{metricgray}\textbf{SA-CER}}\\
Whisp.       & 5.43 & 18.89 & \best{7.74} & 5.36 & 8.38 & 9.34 \\
Kotoba       & 7.16 & 17.97 & 9.66 & 6.59 & 9.62 & 10.27 \\
Rz-NeMo      & 6.00 & 18.36 & 11.03 & 6.66 & 9.67 & 10.17 \\
Rz-k2        & \best{5.02} & 15.77 & 9.85 & 5.20 & 7.96 & 8.51 \\
Qwen         & 6.86 & 13.37 & 9.54 & 6.28 & 7.88 & 8.59 \\
\midrule
Ruby-ASR-ver & 5.95 & 6.55 & 8.67 & 4.78 & \best{7.69} & \best{6.59} \\
Ruby-ASR-sub & 6.03 & \best{6.34} & 8.80 & \best{4.65} & 7.95 & 6.64 \\
\bottomrule
\end{tabular}
\vspace{-15pt}
\end{table}

% -------------------------------------------------------------------------
\section{Results}
\label{sec:results}

\subsection{Orthographic and script-aware recognition}
\vspace{-5pt}

Table~\ref{tab:orth_results} shows no tradeoff in readable transcription:
Ruby-ASR-sub achieves the lowest weighted Raw CER (8.51\%), and Ruby-ASR-ver
the lowest weighted SA-CER (6.59\%), compared with 10.78\% and 8.59\% for
Qwen3-ASR. On JSUT-Book, Ruby-ASR-sub decreases from 23.21\% Raw CER to 8.80\%
SA-CER, indicating that much of the raw error reflects script realization
rather than recognized-reading mismatch.

\begin{table}[t]
\centering
\scriptsize
\setlength{\tabcolsep}{2.6pt}
\renewcommand{\arraystretch}{0.98}
\caption{Kana CER (\%) for lexical-reading transcription.  Ruby-ASR and
Kana-Whisper are scored from direct reading outputs; the remaining
orthographic systems use a shared G2P.  $^\ast$ marks manually verified
reading references.}
\label{tab:kana_results}
\begin{tabular}{@{}lcccccc@{}}
\toprule
Model & B5K$^\ast$ & CSJ$^\ast$ & Book & CV8 & TEDx & W.Avg. \\
\midrule
Whisp.       & 2.22 & 14.63 & \best{4.76} & 2.83 & 6.88 & 6.57 \\
Kotoba       & 3.34 & 13.37 & 6.65 & 3.60 & 7.74 & 7.08 \\
Rz-NeMo      & 1.89 & 12.65 & 5.85 & 3.18 & 6.88 & 6.17 \\
Rz-k2        & 1.78 & 11.61 & 5.29 & 2.71 & 6.31 & 5.64 \\
Qwen         & 3.07 & 9.90 & 6.85 & 3.50 & 6.14 & 5.74 \\
K-Whisp.     & \best{0.87} & \best{4.39} & 5.39 & 5.27 & 7.31 & 4.66 \\
\midrule
Ruby-ASR-CTC & 1.95 & 11.19 & 5.72 & 9.06 & 8.26 & 7.31 \\
Ruby-ASR-ver & 1.08 & 5.41 & 5.70 & \best{2.57} & \best{5.09} & \best{3.75} \\
Ruby-ASR-sub & 1.32 & 5.37 & 5.84 & 2.63 & 5.66 & 4.01 \\
\bottomrule
\end{tabular}
\vspace{-10pt}
\end{table}

\subsection{Lexical-reading recognition}
\vspace{-5pt}

Ruby-ASR-ver and Ruby-ASR-sub obtain weighted Kana CERs of 3.75\% and 4.01\%,
compared with 5.74\% for Qwen3-ASR, 5.64\% for Rz-k2, and 4.66\% for
Kana-Whisper. Ruby-ASR-ver outperforms the standalone mora-CTC path on every
benchmark, reducing weighted error from 7.31\% to 3.75\%, consistent with
treating CTC as auxiliary monotonic supervision rather than a replacement for
contextualized Ruby generation. Although direct Ruby prediction performs best
in aggregate, Kana-Whisper remains best on B5K and CSJ, and Whisper+G2P on
JSUT-Book. These results do not establish whether the AR decoder uses speech
evidence beyond preceding orthographic context; answering that question
requires a controlled intervention outside this paper's scope.

\begin{table}[t]
\centering
\scriptsize
\setlength{\tabcolsep}{4.0pt}
\renewcommand{\arraystretch}{1.0}
\caption{Oracle-orthography G2P diagnostic on independently verified B5K and
CSJ references. Values are Kana CER (\%). Oracle and Qwen hypotheses use the
same G2P and normalization; Ruby-ASR uses its direct reading projection.}
\label{tab:oracle_g2p}
\begin{tabular}{@{}lccc@{}}
\toprule
Condition & Predictor input & B5K & CSJ \\
\midrule
Oracle orth. + G2P & Ground-truth orth. & 1.69 & \best{3.47} \\
Qwen + G2P         & Recognized orth.   & 3.07 & 9.90 \\
Ruby-ASR-ver       & Speech-conditioned & \best{1.08} & 5.41 \\
Ruby-ASR-sub       & Speech-conditioned & 1.32 & 5.37 \\
\bottomrule
\end{tabular}
\vspace{-10pt}
\end{table}

\subsection{Oracle-orthography G2P diagnostic}
\label{sec:oracle_g2p} 
\vspace{-5pt}
Applying the same G2P to ground-truth orthography separates orthographic
recognition errors from text-only reading errors. Oracle G2P in Table~\ref{tab:oracle_g2p} retains 1.69\%
CER on B5K and 3.47\% on CSJ, confirming a nonzero text-only reconstruction
gap; relative to Qwen+G2P, it reduces error by 1.38 and 6.43 points,
respectively, showing the additional effect of orthographic ASR errors.
Ruby-ASR-ver and Ruby-ASR-sub outperform oracle G2P on B5K by 0.61 and 0.37
points but trail it on CSJ by 1.94 and 1.90 points, making the benefit domain
dependent. This comparison is diagnostic rather than causal because the
systems are unmatched and the Ruby decoder also observes preceding
orthographic context.

\begin{table}[t]
\centering
\scriptsize
\setlength{\tabcolsep}{2.6pt}
\renewcommand{\arraystretch}{0.98}
\caption{Results on G2P-hard utterances, defined by nonzero reading edit
distance for oracle-orthography G2P. B5K contains 1,124/5,000 (22.5\%) and CSJ
2,579/8,460 (30.5\%) such utterances. Exact is
utterance-level exact-reading accuracy (\%).}
\label{tab:g2p_hard}
\begin{tabular}{@{}lcc|cc@{}}
\toprule
& \multicolumn{2}{c|}{B5K} & \multicolumn{2}{c}{CSJ} \\
Model & Kana CER & Exact & Kana CER & Exact \\
\midrule
Oracle orth. + G2P & 6.07 & 0.0 & \best{7.65} & 0.0 \\
Qwen + G2P         & 6.13 & 17.6 & 12.84 & 5.4 \\
Rz-k2 + G2P        & 4.79 & 22.8 & 14.44 & 3.6 \\
\midrule
Ruby-ASR-ver        & \best{1.59} & \best{57.9} & 8.02 & 16.8 \\
Ruby-ASR-sub        & 2.24 & 50.3 & 8.02 & \best{17.1} \\
\bottomrule
\end{tabular}
\vspace{-10pt}
\end{table}

\subsection{G2P-hard reading analysis}
\label{sec:g2p_hard}
\vspace{-5pt}

Table~\ref{tab:g2p_hard} localizes gains to readings that the selected G2P
cannot reconstruct from correct orthography. On B5K, Ruby-ASR-ver reduces Kana
CER from 6.07\% for oracle G2P and 4.79\% for Rz-k2+G2P to 1.59\%, with 57.9\%
exact-reading accuracy. On CSJ, oracle G2P has slightly lower CER than Ruby-ASR
(7.65\% vs. 8.02\%), but the Ruby models more often recover the complete
utterance reading (16.8--17.1\% vs. at most 5.4\% for ASR--G2P). Direct
prediction therefore benefits B5K most clearly, whereas spontaneous CSJ
retains a speech-recognition bottleneck. Because this subset depends on one
G2P implementation, it is a targeted diagnostic rather than an intrinsic
partition of reading difficulty.

\begin{table}[t]
\centering
\scriptsize
\setlength{\tabcolsep}{2.7pt}
\renewcommand{\arraystretch}{0.98}
\caption{Conditional reading results on correctly recognized orthographic
spans. Each entry is C-KanaCER / exact-reading accuracy (\%). Ranges cover
Ruby-ASR-ver and Ruby-ASR-sub. The all-span counts are approximately 22.5k for
B5K and 19.3k for CSJ; multi-reading spans comprise about 49\%.}
\label{tab:conditional_reading}
\begin{tabular}{@{}lcc@{}}
\toprule
Subset & B5K & CSJ \\
\midrule
All correct orth.       & 1.71--1.94 / 97.4 & 1.47--1.61 / 97.2 \\
Multi-reading           & 2.8--3.7 / 96.8   & 2.3--2.8 / $\sim$97 \\
Rare ($1\leq f<10$)     & 5.3--5.5 / 86     & 30--32 / 23 \\
Unseen ($f=0$)          & 13.0--13.2 / 70   & 30--31 / 24 \\
\bottomrule
\end{tabular}
\vspace{-10pt}
\end{table}

\subsection{Conditional and ambiguity-sensitive reading accuracy}
\label{sec:conditional_results}
\vspace{-5pt}

Because overall Kana CER is dominated by text-predictable readings, we use
Eq.~\eqref{eq:conditional_cer} to evaluate correctly recognized orthographic
spans. 
Table~\ref{tab:conditional_reading} shows that both Ruby variants exactly
recover about 97\% of readings when the orthographic span is correct.
Multi-reading spans, roughly half of the eligible spans, retain similar
accuracy, suggesting that dictionary-level ambiguity is not the dominant
residual error. Degradation on rare and unseen pairs instead identifies
lexical coverage as the main remaining challenge, especially on spontaneous
CSJ. Because its rare and unseen subsets contain only 245 and 116 spans, these
results are diagnostic. Conditioning on correct orthography prevents
surface-recognition errors from being attributed to reading selection.

\begin{table}[t]
\centering
\footnotesize
\setlength{\tabcolsep}{3.3pt}
\renewcommand{\arraystretch}{1.0}
\caption{Boundary overflow on the filtered ReazonSpeech test subset
($n=3{,}431$). Start and End are normalized insertion rates (\%) before and
after the annotated subtitle span; Total is their sum. Utt. and $\geq$3 count
utterances with any and at least three overflow characters, respectively; Max
is the maximum overflow length. Lower is better.}
\label{tab:reazon_subtitle_analysis}
\begin{tabular}{@{}lcccccc@{}}
\toprule
Model & Start & End & Total & Utt. & $\geq$3 & Max \\
\midrule
Whisp.       & 0.40 & 0.46 & 0.86 & 239 & 111 & 19 \\
Kotoba       & \textbf{0.17} & 0.21 & \textbf{0.38}
             & \textbf{145} & \textbf{45} & \textbf{11} \\
Rz-NeMo      & 0.73 & 0.22 & 0.95 & 296 & 123 & 19 \\
Rz-k2        & 0.78 & \textbf{0.20} & 0.99 & 338 & 127 & 13 \\
\rowcolor{gray!12}
Ruby-ASR-sub & 0.36 & 0.21 & 0.57 & 160 & 61 & 21 \\
\bottomrule
\end{tabular}
\vspace{-10pt}
\end{table}

\subsection{Subtitle-boundary diagnostic}
\label{subsec:subtitle_diagnostic}
\vspace{-5pt}

Ruby-ASR-sub targets concise subtitle-style transcriptions restricted to the
current segment. After numeral normalization and punctuation removal, we align
each hypothesis and reference at the character level. Insertions before the
first or after the last reference character define \emph{start} and \emph{end
overflow}, respectively, and usually reflect neighboring speech captured near
a segment boundary. Thus, boundary overflow measures segment-level output
control rather than hallucination or recognition accuracy. 
Ruby-ASR-sub achieves the second-lowest total overflow (0.57\%) and affects
fewer utterances than Whisper, Rz-NeMo, and Rz-k2. Kotoba-v2 performs best,
consistent with its ReazonSpeech fine-tuning and closer match to the reference
conventions. Despite one 21-character outlier, Ruby-ASR-sub produces
substantially fewer utterances with at least three overflow characters than
the other non-Kotoba systems. Both ReazonSpeech baselines show more start
overflow, suggesting greater sensitivity to preceding-segment speech.
Overall, subtitle-style training improves boundary control. We treat this only
as a boundary diagnostic because the references are insufficiently reliable
for content-level scoring.

% -------------------------------------------------------------------------
\section{Conclusion}
\label{sec:conclusion}

We presented Ruby-ASR, an evidence-preserving refinement of Japanese ASR
supervision. Conventional orthographic targets can collapse different realized
readings into one label, leaving post-hoc G2P to infer the reading from text
alone; span-bound ruby targets instead retain orthographic identity and lexical
reading in one structured sequence. Across
five benchmarks, Ruby-ASR improves direct lexical-reading recognition while
maintaining strong orthographic and script-aware transcription; a separate
diagnostic shows that Ruby-ASR-sub generally restricts output to the annotated
subtitle span. Oracle-orthography, G2P-hard, and conditional-reading analyses
show that direct prediction is particularly effective on B5K cases where the
selected G2P fails, while rare and unseen orthography--reading pairs remain a
major limitation. The released models provide structured Ruby generation and
optional decoder-free mora output for Japanese speech and reading-aware
research.

% -------------------------------------------------------------------------
\footnotesize
\bibliographystyle{IEEEbib}
\bibliography{refs}

\end{document}